\documentclass[a4paper, 10pt, conference]{ieeeconf}      

\IEEEoverridecommandlockouts                              

\usepackage{graphics} 
\usepackage{graphicx}%
\usepackage{epsfig} 
\usepackage{mathptmx} 
\usepackage{mathrsfs}
\usepackage{times} 
\usepackage{amsmath} 
\usepackage{amssymb}  
\usepackage{amsfonts,bm}
\usepackage{multirow}%
\usepackage{listings}%
\usepackage{amssymb}%
\usepackage{pifont}%
\usepackage{textcomp}%
\usepackage{manyfoot}%
\usepackage{booktabs}%
\usepackage{authblk}
\usepackage{soul, color, xcolor}

\usepackage{cite}
\usepackage{adjustbox}

\def\kk{\mathbf{K}}
\def\vx{{\bm{x}}}
\def\vk{{\bm{k}}}

\title{\LARGE \bf
Adaptive Query Prompting for Multi-Domain Landmark Detection
}

\author[1]{Yuhui Li}
\author[2]{Qiusen Wei}
\author[1*]{Guoheng Huang\thanks{* Corresponding author.}}
\author[3]{Xiaochen Yuan}
\author[4]{Guo Zhong}
\author[5]{Xuhang Chen}

\affil[1]{School of Computer Science and Technology, Guangdong University of Technology, Guangzhou, China}
\affil[2]{School of Automation, Guangdong University of Technology, Guangzhou, China}
\affil[3]{Faculty of Applied Sciences, Macao Polytechnic University, Macao, China}
\affil[4]{School of Information Science and Technology, Guangdong University of Foreign Studies, Guangzhou, China}
\affil[5]{School of Computer Science and Engineering, Huizhou University, Huizhou, China}

\begin{document}

\maketitle
\thispagestyle{empty}
\pagestyle{empty}

\begin{abstract}
Medical landmark detection is crucial in various medical imaging modalities and procedures. Although deep learning-based methods have achieve promising performance, they are mostly designed for specific anatomical regions or tasks. In this work, we propose a universal model for multi-domain landmark detection by leveraging transformer architecture and developing a prompting component, named as Adaptive Query Prompting (AQP). Transformer backbone architecture is suitable for our study due to its ability to capture long-range dependencies crucial in multi-domain landmark detection. Specifically, transformers excel at understanding global anatomical relationships, which span across entire images. Instead of embedding additional modules in the backbone network, we design a separate module to generate prompts that can be effectively extended to any other transformer network. In our proposed AQP, prompts are learnable parameters maintained in a memory space called prompt pool. The central idea is to keep the backbone frozen and then optimize prompts to instruct the model inference process. Furthermore, we employ a lightweight decoder to decode landmarks from the extracted features, namely Light-MLD. Thanks to the lightweight nature of the decoder and AQP, we can handle multiple datasets by sharing the backbone encoder and then only perform partial parameter tuning without incurring much additional cost. It has the potential to be extended to more landmark detection tasks. We conduct experiments on three widely used X-ray datasets for different medical landmark detection tasks. Our proposed Light-MLD coupled with AQP achieves SOTA performance on many metrics even without the use of elaborate structural designs or complex frameworks. 

\end{abstract}

\section{INTRODUCTION}

Landmark detection plays an important role in various of medical image analysis tasks \cite{Chen2024-dg,Gong2023GenerativeAF,Huang2023MRIS,chen11,liu2020fine,li1,li2}. It aims to localize anatomical keypoints in medical images such as X-ray images. Manual annotation is highly time-consuming, labor-intensive and typically requires a high level of expertise. Automatic annotation methods can significantly reduce the required labor and enable the analysis of large-scale datasets \cite{medsam}. 
Despite the fact that deep learning-based methods have shown great promise on in the field of medical landmark detection \cite{chen12,chen13,chen14,jiang2021deep,jiang2020geometry,zhang2022correction,li2020diagnosis,li2021hybrid,zhang1,li2023optimized,yang2024adaptive,huang2024deformmlp,li2022few,zhang2,zhang3,zhang4,li2022monocular,song2024local,li2023cee,zhang5,zhang6,zhang7,zhang8,zhang9,zhang10}, a significant limitation of many existing models is their task-specific nature. These models are typically tailored and refined to optimize performance within the confines of their designated tasks. Zhu et al. propose GU2Net \cite{yolo} and DATR \cite{datr} as universal models for multi-domain landmark detection. However, these methods focus on elaborate structural designs to obtain better model performance, and require to train the entire model when handling new tasks, which is highly inefficient. 

\begin{figure}[t]
    \centering
    \includegraphics[width=\linewidth]{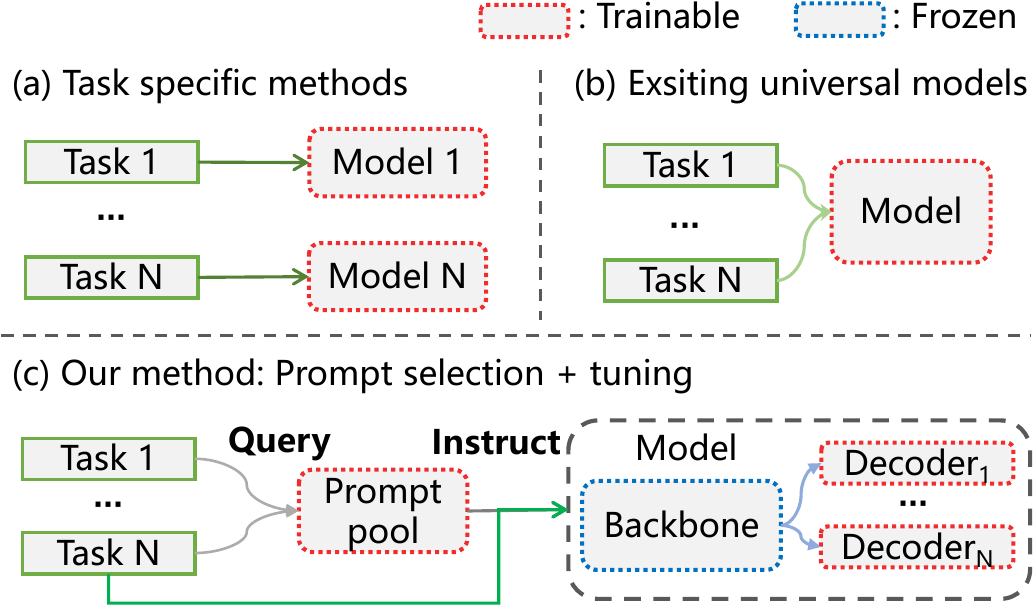}
    \caption{Overview of the AQP framework. Compared with typical task-specific methods and existing universal models, which adapt entire model weights to deal with new tasks, AQP uses a single frozen backbone model and learns a prompt pool to instruct the model conditionally. }
    \label{fig:comparision}
\end{figure}

To this end, we propose \emph{Adaptive Query Prompting} (AQP), a novel prompting method for muti-domain learning. Figure~\ref{fig:comparision} gives an overview of our method in contrast to typical task specific methods and existing universal models. The basic idea is to efficiently adapt a pre-trained large foundation model to many downstream tasks with the minimum extra trainable parameters \cite{evp,jia2022visual,liu2023explicit2}. In the context of our multi-domain landmark detection task, AQP allows us to use the same transformer backbone for head, hand, and chest X-ray landmark detection by learning task-specific prompts. This is in contrast to conventional approaches that might require training separate models for each task or fine-tuning the entire model for each domain. We keep the pre-trained model untouched, and instead learn a set of prompts that dynamically instruct model to solve corresponding tasks, it leverages the representative features from pre-trained models. The prompts are selected via a query-based mechanism. Specifically, prompts and keys are paired in a shared memory space called prompt pool, then we design a query function to get proper prompts paired with the keys, which are selected based on the query function. Our proposed AQP can select the appropriate prompt for various input, and does not require complex design. The prompt pool, optimized jointly with the supervised loss, serves to ensure that shared prompts encode shared knowledge for effective knowledge transfer, while unshared prompts encode task-specific knowledge crucial for preserving model plasticity. The trainable parameters of our method is much fewer, thus being able to adapt to different tasks more efficiently.

In order to handle multiple landmark detection datasets without excessive extra training cost, we employ a plain vision transformer as backbone to encode features and then adopt multiple lightweight decoders to deal with different tasks~\cite{Chen2023ALF,Li2023HighResolutionDS,Luo2023DevignetHV,liu2023coordfill,chen15}. For each iteration, we randomly sample instances from multiple training datasets and feed them into the backbone and the decoders to estimate the heatmaps corresponding to each dataset. Compared to previous methods, AQP enjoys the structure simplicity and can adapt well to different tasks through fine-tuning.

In summary, our  main contributions can be outlined as follows:
\begin{enumerate}

\item We propose \emph{Adaptive Query Prompting} (AQP), a novel prompt tuning method for multi-domain learning. It leads to better model generalization on the downstream tasks. We conduct the ablation to show the effectiveness of the prompting components.

\item We propose a unified transformer network for multi-domain landmark detection, namely Light-MLD, following the encoder-decoder pipeline. With the lightweight decoder, we can handle new tasks without incurring much additional cost.

\item  In terms of experiments, we validate our method on three datasets for different medical landmark detection tasks: head, hand and chest. Our propose AQP achieves competitive performance with the whole model training and task-specific solutions without modification. Our model is simple but even outperforms previous state-of-the-art methods on many metrics.

\end{enumerate}

\section{RELATED WORK}

\subsection{Vision Transformer for Medical Landmark Detection}

Recently, vision transformers \cite{vit,vitpose} have shown great potential in many vision tasks~\cite{Chen2022ShadocnetLS,li2022wavenhancer,liu2024depth,chen10,chen7,liu2024dh}. For instance, Xiao et al. \cite{bib3} propose a lightweight Transformer-embedded network for vertebrae landmark detection, which has yielded promising results in this task. PRTR \cite{bib41} incorporates both transformer encoders and decoders to gradually refine the locations of the estimated keypoints in a cascade manner. However, Most of these approaches employ a convolutional neural network (CNN) as a backbone, followed by the integration of a transformer with intricate architectures to enhance the extracted features and capture the relationships among the keypoints. They either necessitate CNNs for feature extraction or demand meticulous design of transformer architectures. In the contrast, ViTPose \cite{vitpose} employs a plain vision transformer \cite{vit} as the backbone with a simple decoder and obtains SOTA performance on representative benchmarks without elaborate designs. Follow its success, we introduced this framework into medical landmark detection.

\subsection{Prompt Tuning}
Prompting was originally introduced in the field of natural language processing (NLP) \cite{bib15}. Shyam et al. \cite{bib6} illustrates robust generalization to downstream transfer learning tasks, even in few-shot or zero-shot settings, utilizing manually selected prompts within GPT-3. Recently, the concept of prompting \cite{vpt,bib17} has been extended to vision tasks. Sandler et al. \cite{bib17} introduces memory tokens, which are sets of trainable embedding vectors for each transformer layer. These traditional prompt tuning methods design task-specific prompt functions to instruct pre-trained models perform corresponding tasks conditionally \cite{bib15,Chen2023MedPromptCP}, so that the language model gets additional information about the task. However, crafting an effective prompting function poses a challenge, which require heuristic strategies. In this work, instead of designing a prompting function, we learn a set of prompts stored in the prompt pool. A query function is used to dynamically select suitable prompts tailored to different inputs. All that remains is the design of an effective query function, which is much easier than design prompts for different tasks. We can utilize the entire pre-trained model as a frozen feature extractor to obtain the query features.

\section{METHOD}

\textbf{Problem Definition:} Let ${D_1,D_2,\dots,D_N}$ denote a set of $N$ datasets, potentially originating from distinct anatomical regions. Given an image $I_i\in \mathbb{R}^{C_i\times H_i\times W_i}$ from Dataset $D_i$, along with corresponding landmarks ${(x_{iC_i^{'}},y_{iC_i^{'}})}$, the $k^{th}$ ($k \in [1,2,\ldots,C_i^{'}]$) landmark's heatmap $Y_{ik} \in \mathbb{R}^{C_i^{'}\times H_i\times W_i}$ is obtained using the Gaussian function:
\begin{equation}
Y_{ik}= \frac{1}{\sqrt{2\pi}\sigma}\exp\left({-\frac{(x-x_{ik})^2+(y-y_{ik})^2}{2\sigma^2}}\right),
\label{Eq:gt}
\end{equation}
where $C_i$ represents the number of channels in the input image; $C_i^{'}$ denotes the number of channels in the output heatmap, namely the number of landmarks. The goal is to learn the Gaussian distribution (the heatmaps) corresponding to the landmarks.

\begin{figure*}[ht]
    \centering
    \includegraphics[width=0.8\linewidth]{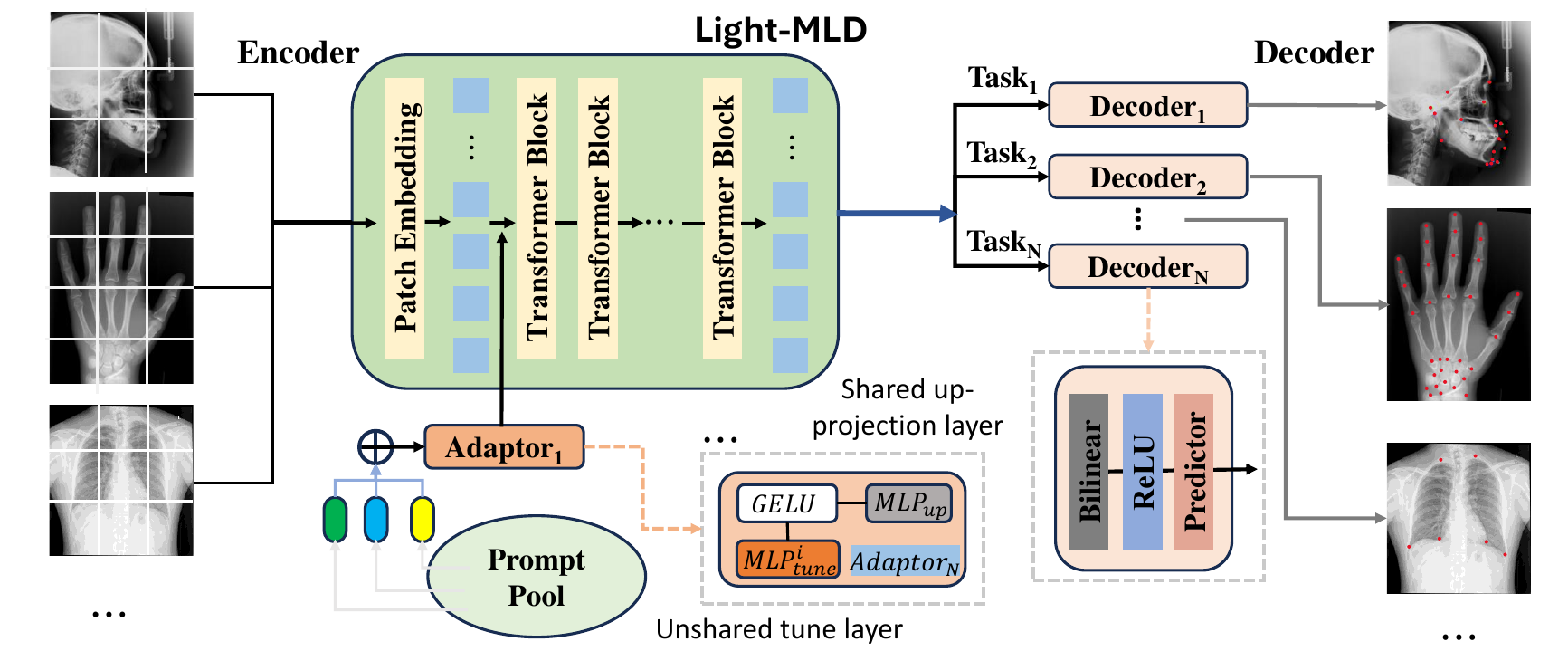}
    \caption{The framework of Light-MLD. The backbone is consist of several feature embedding layers and vision transformer blocks, following by several decoder layers.}
    \label{fig:framework}
\end{figure*}

\subsection{Preliminaries}\label{subsec2}
\subsubsection{ViTPose} 
ViTPose \cite{vitpose} is a transformer-based network for human pose estimation. Similar to traditional CNN backbone \cite{bib18}, ViTPose captures multi-stage features via several stages. Differently, each stage is built via the feature embedding layers and vision transformer blocks \cite{vit,attention}. ViTPose adopt plain vision transformer with masked image modeling pre-training backbones, then pre-training the backbones using MAE \cite{mae} on MS COCO \cite{coco} dataset and a combination of MS COCO and AI Challenger \cite{bib22} respectively by random masking 75\% patches from the images and reconstructing those masked patches. In this work, we use the pre-trained weights to initialize the backbone of our model. 

\subsubsection{Lightweight decoder}
Unlike the existing methods which use elaborate designs to decode the heatmaps from extracted features, we avoid introducing fancy but complex modules to keep the structure as simple as possible. We simply append several decoder layers after the transformer backbone to estimate the heatmaps. As shown in Figure~\ref{fig:framework}, we directly utilize bilinear interpolation to upsample the feature maps by a factor of 4, followed by applying a Rectified Linear Unit (ReLU) and a convolutional layer with a kernel size of $3 \times 3$ to obtain the heatmaps, \emph{i.e.},
\begin{equation}
    K = \operatorname{Conv}_{3\times3}(\operatorname{Bilinear}(\operatorname{ReLu}(F_{out}))),
\end{equation}
where $F_{out}$ is the  output feature of the transformer backbone. Since the simplicity and lightweight nature of the decoder, Light-MLD can effectively handle different tasks by incorporating multiple decoders without incurring much additional cost.

\subsubsection{Prompt-based learning} 
Given an input of 2D image $I\in \mathbb{R}^{H\times W\times C}$ and a pre-trained vision transformer (ViT) backbone $f = f_r \circ f_e$ (excluding the decoder head), where $f_r$ denotes transformer blocks \cite{vit}, and $f_e$ represents the patch embedding layer. Images are reshape to a sequence of flattened 2D patches $I^p \in \mathbb{R}^{N\times(C\cdot S)}$, where $N$ is the number of patches. A image patch $I^p$ is projected to a D-dimension feature through the patch embedding layer, \emph{i.e.}, $f_e  :\mathbb{R}^{N\times(C\cdot S)} \rightarrow \mathbb{R}^{S\times D}$ , where $D$ is the embedding dimension. The extracted features are denoted as $x_e$, and $x_e = f_e(I_p)\in \mathbb{R}^{S\times D}$. When employing prompt tuning methods, prompts can be fed into model in various ways. For example, a prompt $P_e\in\mathbb{R}^{L_p\times D}$ is considered as an additional token concatenated to the embedding future $x_e$, where $L_p$ is the token length. In a remarkable work called Explicit Visual Prompting (EVP) \cite{evp}, prompts are first adjusted by a tunable adapter, then added to the extracted embedding feature $x_e$ through element-wise addition. This kind of prompt tuning typically yields better results because it learns explicit prompts. In this work, we adopt a approach similar to EVP.

\subsection{Overview}
As illustrated in Figure~\ref{fig:framework}, Light-MLD employs a transformer-based backbone as encoder, which captures multi-stage features via several vision transformer blocks, then followed by a series of lightweight decoders to decode the corresponding landmarks for different tasks. Figure~\ref{fig:prompt} gives a overview our proposed \emph{Adaptive Query Prompting} (AQP). Prompts are selected from the prompt pool to instruct the model during training process.

\subsection{Adaptive query prompting}\label{subsec3}
In our proposed adaptive query prompting, there are three main components, namely the Prompt pool, the Prompt query mechanism and the Adaptor. It is worth mentioning that AQP only adds negligible additional parameters ($\sim$ 0.1\%). Although AQP introduces extra prompts, its parameters are remarkably intricate, embodying a fundamentally different design principle from architecture-based methods: AQP designs a novel memory mechanism based on prompts, learning high-level instructions from model inputs to guide model outputs while maintaining the learned architecture unchanged. In contrast, most architecture-based methods aim to separate model parameters.

\subsubsection{Prompt pool} 
The reasons for adopting prompt pooling are twofold. On the one hand, designing an effective prompt generation function is nontrivial, especially for a generic landmark detection model. On the other hand, a shared prompt pool enables knowledge transfer between different tasks, since there are commonalities between different tasks. Thus, we employ a \emph{prompt pool} to store encoded knowledge, which can be flexibly grouped as input to the model. The prompt pool is defined as 
\begin{equation}
\label{eq:prompt_pool}
    \mathbf{P}=\{P_1, P_2, \cdots, P_N\},
\end{equation}
where $N$ represents total number of prompts. $P_j \in \mathbb{R}^{L_p\times D}$ is a single prompt with the same embedding size $D$ as $x_e$ and token length $L_p$. $x_e$ is extracted embedding feature projected by the embedding layer. During the training phase, the prompt pool is randomly initialized.

\subsubsection{Prompt query mechanism} 
\begin{figure}[h]
    \centering
    \includegraphics[width=\linewidth]{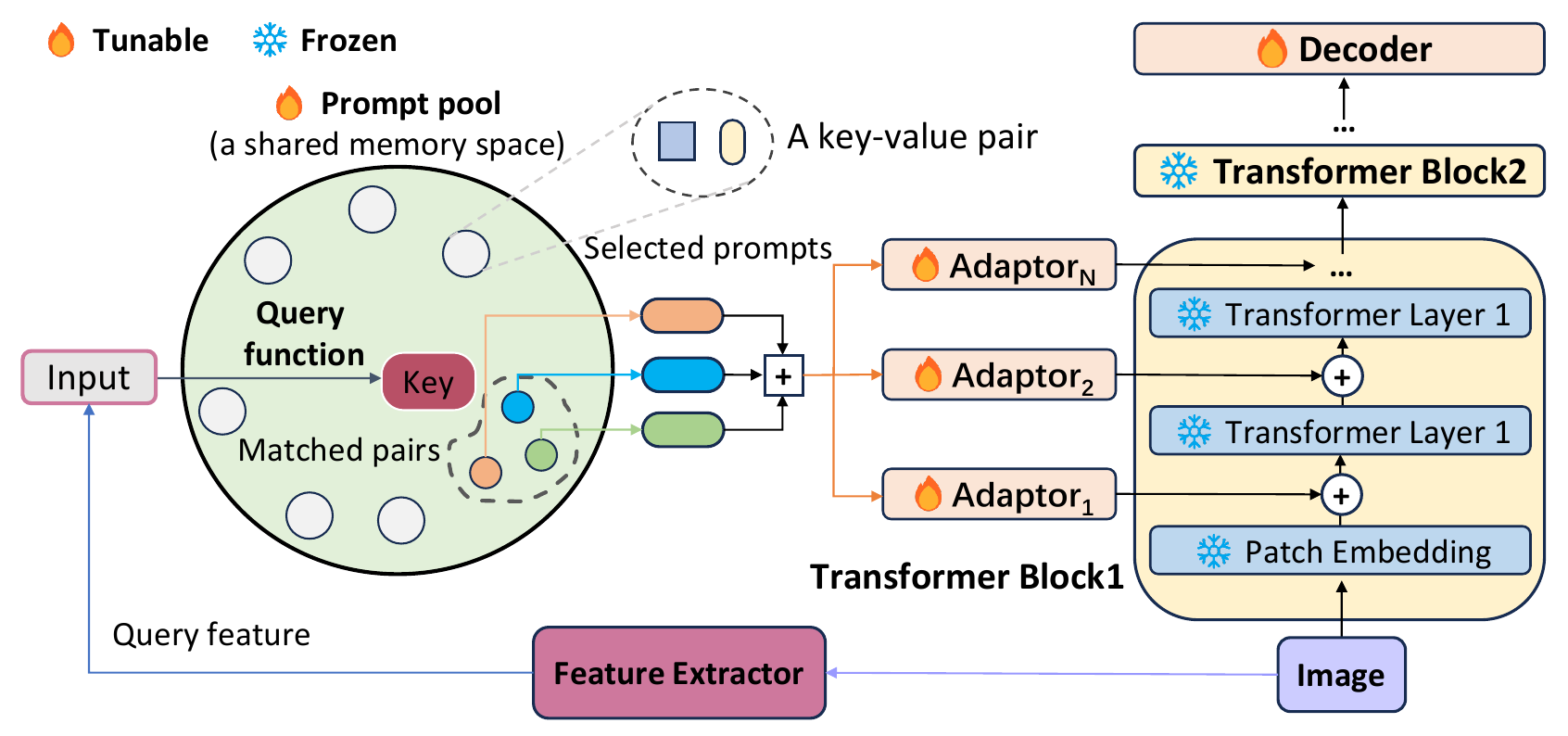}
    \caption{The architecture of the proposed AQP, where the operator $\boxplus$ denotes concatenate and $\bigoplus$ means element-wise addition. The goal of the Adaptor is to merge selected prompts and align them with the input.}
    \label{fig:prompt}
\end{figure}

In order to efficiently select suitable prompts for different inputs, We propose an approach of query strategy based on key-value pairs, which dynamically determines appropriate prompts for varying inputs. As shown in Figure~\ref{fig:prompt}, this query mechanism relies on key-valued memory, which uses external memory for distinct computational objectives. Each prompt is associated with a learnable key in the form of ${(k_1,P_1), (k_2, P_2), \cdots, (k_N, P_N)},$ where $k_i\in \mathbb{R}^{D_k}$. The collection of all keys is denoted as $\mathbf{K}=\{k_i\}^N_{i=1}$. Ideally, we aim to enable the input instance itself to determine the prompts to select via query-key matching. For this purpose, we introduce a query function $q : \mathbb{R}^{H\times W\times C} \rightarrow \mathbb{R}^{D_k}$ that encodes the input $I$ to the same dimension as the keys in the prompt pool. Furthermore, $q$ should be a deterministic function across different tasks and devoid of learnable parameters. In this work, we utilize the pre-trained transformer backbone as a fixed feature extractor to obtain the query features.

Let $\gamma : \mathbb{R}^{D_k} \times \mathbb{R}^{D_k} \rightarrow \mathbb{R}$ represents a function that scores the match between the query and prompt key (in this work we adopt cosine distance).  When presented with an input $x$, we use $q(x)$ to retrieve the top-M keys by straightforwardly solving the objective function:
\begin{equation}
    {\kk_\vx} = \underset{\{s_i\}_{i=1}^{M} \subseteq [1, N] }{\operatorname{argmin}} \quad \sum_{i=1}^{M} \gamma\left({q(\vx), \vk_{s_i}}\right),
\label{eq:lookup}
\end{equation}
where $\kk_\vx$ denotes a subset of top-M keys specifically chosen for $\vx$ from $\kk$. Furthermore, the querying of prompts is performed on an instance-wise basis, rendering the entire framework task-agnostic, which enhances its generality.

\subsubsection{Adaptor}
The objective of the Adaptor is to conduct adaptation efficiently and effectively across all layers, incorporating features from all the prompts selected from the prompt pool. Denote $P$ as the concatenation of $M$ selected prompts, $P =  [P_1, P_2, \cdots, P_M]$. For the $i$-th Adaptor, we take $P$ as input to generate the final prompting $P^i_f$:
\begin{equation}
    P^i_f = \operatorname{MLP_{up}}(\operatorname{GELU}(\operatorname{MLP^i_{tune}}(P))),
\end{equation}
where $MLP_{up}$ represents an up-projection layer shared across all Adaptors, aimed at aligning the dimensions of transformer feature. $GELU$ denotes the GELU activation function \cite{bib53}. $MLP_{tune}$ is a linear layer responsible for generating distinct prompts in each Adaptor. $P^i_f$ is the output prompt that attaches to each transformer layer. This design allows for strong adaptability while generating diverse prompts. With the adaptor, we can generate suitable prompts for all tasks in a unified paradigm even when using different models or backbones.

\subsection{Training}\label{subsec4}

\subsubsection{Prompting details}
Here we elaborate on some details about prompting during training process. Given an image input $\vx$, we first obtain the query features through a feature extractor. In this work, we directly adopt the entire pre-trained model as a frozen feature extractor, other pre-trained model like ConvNet are feasible as well. The corresponding query features is denoted as $q(\vx)$. After obtaining the query features, we derive the corresponding keys by simply solving Equation \ref{eq:lookup}, where $\gamma$ is a function used to measure the degree of matching between query features and keys, namely the query function. $M$ is a hyperparameter determines the number of prompts selected in the prompt pool. Here we use cosine similarity as the measure. The selected prompts are concatenated then fed into the adaptors, which perform adaptation across all the layers, as shown in Figure~\ref{fig:prompt}. Finally, the output prompts are attached to each transformer layer guiding the model's inference process.

\subsubsection{Loss function}
In this work, we jointly train the two components, the original model and the prompt pool. Overall, we seek to minimize the following training loss function:
\begin{equation} \label{eq:full_loss}
\begin{split}
\mathcal{L} = \frac{1}{2N} \sum_{j=0}^{N-1}||H^j - G^j||_{2} + \lambda \sum_{\kk_{\vx}} \gamma\left({q(\vx), \vk_{s_i}}\right), 
\end{split}
\end{equation}
where $\kk_{\vx}$ is obtained with Equation~\eqref{eq:lookup}.

The first term is the heatmap MSE loss $\mathcal{L}_{ht}$, where $H^j$ represents the predicted heatmap for the $j$-th keypoint, $G^j$ denotes the corresponding ground truth gaussian heatmap for the $j$-th keypoint. $N$ is the total number of keypoints, and $||\cdot||_{2}$ denotes the L2 norm between the predicted and ground truth heatmaps. The second term serves as a surrogate loss to bring the selected keys closer to their corresponding query features, and $\lambda$ is a scalar used to weight the loss.

\section{EXPERIMENTS}\label{sec4}
In this section, we conduct quantitative and qualitative evaluations of our universal model, comparing it with SOTA methods on three public X-ray datasets of head, hand and chest for different medical landmark detection tasks. Moreover, we conduct extensive ablation studies to demonstrate the effectiveness of the proposed AQP.

\subsection{Datasets}\label{subsec41}
\subsubsection{Head}
The head dataset is a widely-used public dataset for cephalometric landmark detection, which contains 400 X-ray images, provided in IEEE ISBI 2015 challenge \cite{bib42}. It contains 19 landmarks for each X-ray image. For our experimentation, we designate the first 150 images for training and the remaining 250 images for testing purposes.

\subsubsection{Hand}
The hand dataset comprises 909 X-ray images. We allocate the first 609 images for training and reserve the remaining 300 images for testing. Payer et al. \cite{bib43} manually labeled a total of 37 landmarks.

\subsubsection{Chest}
The chest dataset comprises 279 X-ray images. The initial 229 images are allocated for training, and the remaining 50 images are reserved for testing purposes. There are six landmarks labeled in each chest X-ray image, which delineate the boundary of the lung.

\begin{table*}[ht]
\centering
\caption{Quality metrics of different models on head and hand datasets. + represents the model is learned on the mixed datasets. The best results are in \textbf{bold} and the second best results are \underline{underlined}. }
\label{tab_results}
\adjustbox{width=\linewidth}{
\begin{tabular}{lccccccccccccc}
\toprule
\multirow{2}*{Models}  &\multirow{2}*{MRE} &\multicolumn{4}{c}{Head  SDR(\%)}   &\multirow{2}*{MRE} & \multicolumn{3}{c}{Hand SDR(\%)}  & \multirow{2}*{MRE} &  \multicolumn{3}{c}{Chest SDR(\%)}\\

\cmidrule(l){3-6}\cmidrule(l){8-10}\cmidrule(l){8-10}\cmidrule(l){12-14}   &(mm)& 2mm   & 2.5mm & 3mm   & 4mm   &(mm)     & 2mm & 4mm & 10mm  &(px)  & 3px & 6px & 9px\\
\midrule
\multirow{1}*{Ibragimov et al.~\cite{bib47}}  &1.84& 68.13 & 74.63 & 79.77 & 86.87  &-    &-&-&-    &-&-&-&-     \\
\multirow{1}*{{\v{S}}tern et al.~\cite{bib52}}    &-&-&-&-&-      &{0.80} & 92.20 & 98.45 & 99.83   &-&-&-&- \\
\multirow{1}*{Lindner et al.~\cite{bib48}}      &{1.67}& 70.65 & 76.93 & 82.17 & {89.85} &0.85 & 93.68 & 98.95 & 99.94 &-&-&-&-\\
\multirow{1}*{Urschler et al.~\cite{bib49}}    &-& 70.21 & 76.95 & 82.08 & 89.01     &{0.80}  & 92.19 & 98.46 & {99.95} &-&-&-&-\\
\multirow{1}*{Payer et al.~\cite{bib50}}            &-& {73.33} & {78.76} & {83.24} & 89.75   &\textbf{0.66} & {94.99} & {99.27} & \textbf{99.99} &-&-&-&-\\
\multirow{1}*{U-Net~\cite{unet}+}                  &12.45& 52.08 & 60.04 & 66.54 & 73.68    &6.14  & 81.16 & 92.46 & 93.76 &{5.64} &{51.67} & {82.33} & \textbf{90.67}\\ 
\multirow{1}*{GU2Net~\cite{yolo}+}                        &\underline{1.56}& \textbf{77.79} & \textbf{84.65} & \underline{89.41} & \underline{94.93}   &0.84& \underline{95.40} & \underline{99.35} & {99.75} &\textbf{5.57}& \textbf{57.33} &82.67 & {89.33}\\
\multirow{1}*{AQP (Ours)+}                        &\textbf{1.54}& \underline{76.39} & \underline{84.43} & \textbf{89.61} & \textbf{95.02}   &\underline{0.76}& \textbf{96.27} & \textbf{99.45} & \underline{99.79} &\underline{5.61} &\underline{56.78} &\textbf{83.46} & \underline{90.04}\\
\bottomrule
\end{tabular}}
\end{table*}

\begin{figure*}[h]
    \centering
    \includegraphics[width=0.8\linewidth]{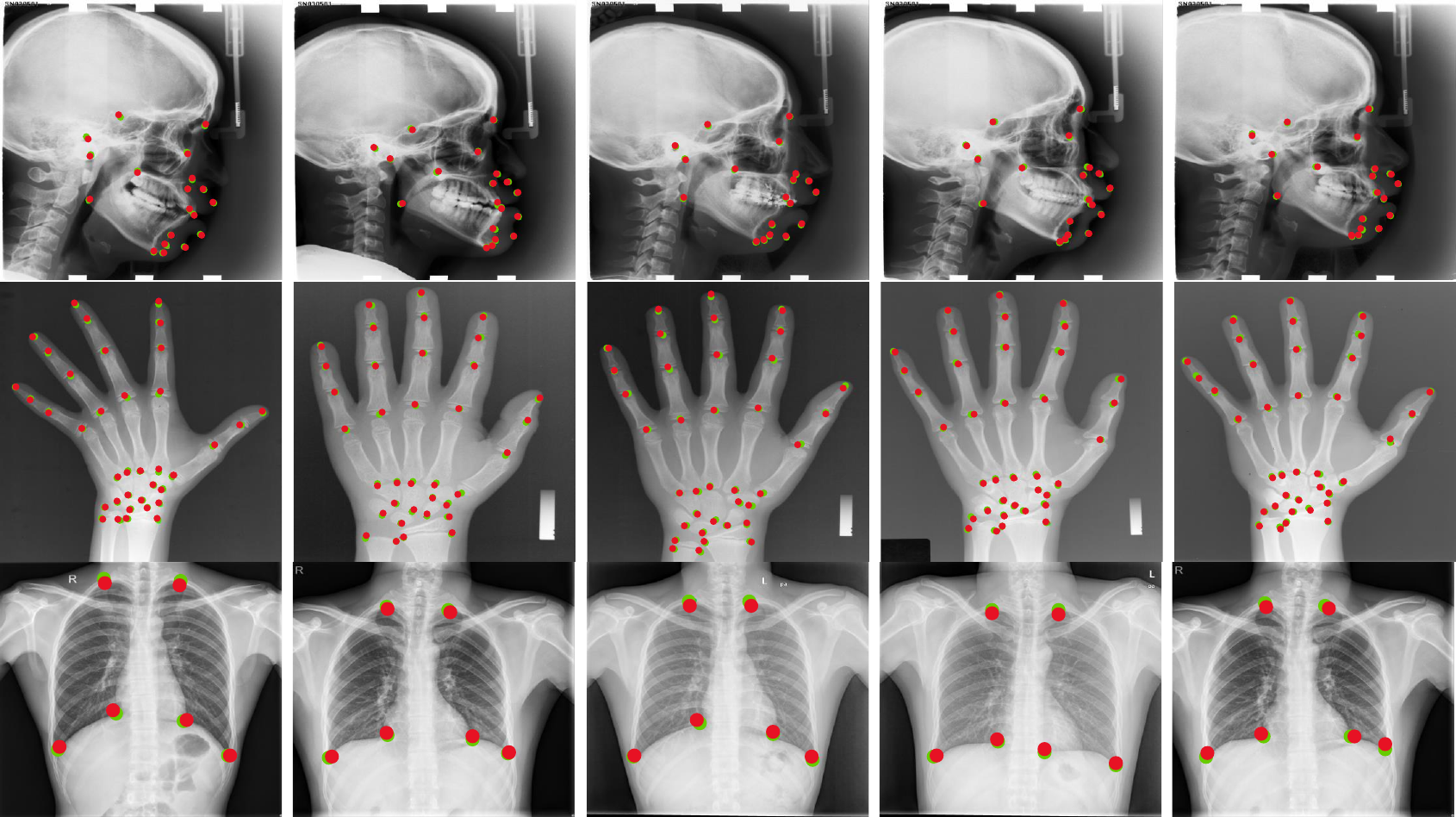}
    \caption{Subjective results of the head, hand and chest datasets. All images are randomly selected. The red points \textcolor{red}{$\bullet$} are the landmarks predicted by our model while the green points \textcolor{green}{$\bullet$} are the ground truth labels.}
    \label{fig:subjective}
\end{figure*}

\subsection{Settings}\label{subsec42}
All experiments are conducted on a single NVIDIA A100 GPU with 40G of memory. For model training, we adopt the AdamW optimizer with an initial learning rate of $\text{5e}^{-4}$. Cosine decay is employed to adjust the learning rate over time. We use the ViTPose-L as backbones for test, which are pre-trained on MS COCO Keypoint Detection Dataset \cite{coco}, corresponding to ViT-Large. The models are trained 25 epochs for all datasets. The images are resized to the shape of $256\times 192$ for all datasets.

\subsection{Evaluation and results}\label{subsec43}
For the evaluation metric, we use the mean radial error (MRE) to measure the Euclidean distance between prediction and ground truth, and the sccessful detection rate (SDR).

Table \ref{tab_results} presents the experimental results on the the head, hand and chest datasets. Here we adopt the ViTPose-L as backbone and fine-tune the model with AQP on these datasets. From the experimental results, it can be observed that our Light-MLD with AQP achieves performance comparable to the previous SOTA methods. On the head dataset, Light-MLD achieved the best accuracy of 89.61\% within 3mm and 95.62\% within 4mm, and second-best performance on the remaining metrics, falling just slightly behind the best-performing model. On the hand dataset, Light-MLD outperforms the previous SOTA method for SDR within 2mm and 3mm. On the chest dataset, our model also reaches the best accuracy of 83.46\% with 6px, but a little worse on the other metrics. Overall, our model achieve the best or second-best performance on the three datasets across all metrics, and performs better than models that are learned on a single dataset.
We also provide some visual landmark detection results for subjective evaluation, as shown in Figure~\ref{fig:subjective}. We demonstrate the results on head, hand and chest datasets respectively.

\subsection{Ablation Study}\label{subsec44}
To evaluate the effectiveness of the proposed Adaptive Query Prompting (AQP) method, we conduct ablation experiments on the head dataset using different ViTPose backbones. We compare the performance of models fine-tuned with and without AQP, denoted by ``\#'' in Table~\ref{abs}.

As shown in Table~\ref{abs}, incorporating AQP consistently improves performance across all backbones and evaluation metrics. This highlights the effectiveness of AQP in adapting a pre-trained model for multi-domain landmark detection. Specifically:
\begin{itemize}
    \item Improved MRE: AQP consistently reduces the MRE across all backbones, indicating more accurate landmark localization.
    \item Enhanced SDR: AQP significantly improves the SDR, especially at stricter thresholds (2mm, 2.5mm), demonstrating its ability to achieve higher detection accuracy.
\end{itemize}
These results demonstrate that AQP effectively leverages the pre-trained knowledge from different backbones and adapts it to the specific task of multi-domain landmark detection, leading to improved performance compared to models without AQP.

\begin{table}[!ht]
\centering
\caption{Ablation study of our universal model with prompt tuning method using different backbones. $\#$ denotes the models are fine-tuned with AQP.}
\label{abs}
\begin{tabular}{l|ccccc}
\toprule
\multicolumn{1}{c}{\multirow{2}{*}{Model}} & \multirow{2}{*}{\begin{tabular}[c]{@{}c@{}}MRE($\downarrow$)\\ mm\end{tabular}} & \multicolumn{4}{c}{SDR($\uparrow$)(\%)}       \\ 
\multicolumn{1}{c}{}                       &                                                                   & 2mm   & 2.5mm & 3mm   & 4mm   \\  \midrule 

ViTPose-S                           & 1.83 & 67.31 & 74.36 & 79.71 & 86.52 \\
ViTPose-S\#                        & \textbf{1.73} & \textbf{69.89} & \textbf{77.03} & \textbf{82.01} & \textbf{89.55} \\
ViTPose-B                           & 1.76 & 70.24 & 76.59 & 82.17 & 89.77 \\
ViTPose-B\#                        & \textbf{1.63} & \textbf{74.89} & \textbf{80.23} & \textbf{86.91} & \textbf{92.03} \\
ViTPose-L                           & 1.62 & 75.81 & 80.71 & 86.54 & 92.68 \\
ViTPose-L\#                         & \textbf{1.56} & \textbf{76.39} & \textbf{83.43} & \textbf{89.61} & \textbf{95.02} \\

\bottomrule 
\end{tabular}
\end{table}

\section{CONCLUSIONS}
This paper presents Adaptive Query Prompting (AQP) for Multi-Domain Landmark Detection. Our universal model Light-MLD is simple yet powerful, consisting only of a plain vision transformer backbone and a simple decoder. Furthermore, the proposed AQP can effectively instruct the model perform better without incurring much additional cost. Experimental results demonstrate that our proposed method performs well in different landmark detection tasks, even outperforms previous state-of-the-art methods on many metrics. For future work, we will derive more complex framework to test the effectiveness of our AQP, since the model in this work is quite simple.


\bibliographystyle{IEEEtran}
\bibliography{ref}

@article{bib3,
  title={Application of TVD-Net for sagittal alignment and instability measurements in cervical spine radiographs},
  author={Xiao, Qiangqiang and Chen, Yao and Wang, Jianxi and Zang, Fazhi and Wang, Yunhao and Zheng, Genjiang and Yang, Kunyu and Zhang, Rongcheng and Hu, Bo and Chen, Huajiang},
  journal={Medical Physics},
  year={2023},
  publisher={Wiley Online Library}
}

@article{medsam,
  title={Segment anything in medical images},
  author={Ma, Jun and He, Yuting and Li, Feifei and Han, Lin and You, Chenyu and Wang, Bo},
  journal={Nature Communications},
  volume={15},
  number={1},
  pages={654},
  year={2024}
}

@article{bib6,
  title={Language models are few-shot learners},
  author={Brown, Tom and Mann, Benjamin and Ryder, Nick and Subbiah, Melanie and Kaplan, Jared D and Dhariwal, Prafulla and Neelakantan, Arvind and Shyam, Pranav and Sastry, Girish and Askell, Amanda and others},
  journal={NeurIPS},
  volume={33},
  pages={1877--1901},
  year={2020}
}

@article{vitpose,
  title={Vitpose: Simple vision transformer baselines for human pose estimation},
  author={Xu, Yufei and Zhang, Jing and Zhang, Qiming and Tao, Dacheng},
  journal={NeurIPS},
  volume={35},
  pages={38571--38584},
  year={2022}
}

@inproceedings{evp,
  title={Explicit visual prompting for low-level structure segmentations},
  author={Liu, Weihuang and Shen, Xi and Pun, Chi-Man and Cun, Xiaodong},
  booktitle={CVPR},
  pages={19434--19445},
  year={2023}
}

@article{vit,
  title={An image is worth 16x16 words: Transformers for image recognition at scale},
  author={Dosovitskiy, Alexey and Beyer, Lucas and Kolesnikov, Alexander and Weissenborn, Dirk and Zhai, Xiaohua and Unterthiner, Thomas and Dehghani, Mostafa and Minderer, Matthias and Heigold, Georg and Gelly, Sylvain and others},
  journal={arXiv},
  year={2020}
}

@article{bib15,
  title={Pre-train, prompt, and predict: A systematic survey of prompting methods in natural language processing},
  author={Liu, Pengfei and Yuan, Weizhe and Fu, Jinlan and Jiang, Zhengbao and Hayashi, Hiroaki and Neubig, Graham},
  journal={ACM Computing Surveys},
  volume={55},
  number={9},
  pages={1--35},
  year={2023},
  publisher={ACM New York, NY}
}

@inproceedings{vpt,
  title={Visual prompt tuning},
  author={Jia, Menglin and Tang, Luming and Chen, Bor-Chun and Cardie, Claire and Belongie, Serge and Hariharan, Bharath and Lim, Ser-Nam},
  booktitle={ECCV},
  pages={709--727},
  year={2022},
  organization={Springer}
}

@inproceedings{bib17,
  title={Fine-tuning image transformers using learnable memory},
  author={Sandler, Mark and Zhmoginov, Andrey and Vladymyrov, Max and Jackson, Andrew},
  booktitle={CVPR},
  pages={12155--12164},
  year={2022}
}

@inproceedings{bib18,
  title={Deep residual learning for image recognition},
  author={He, Kaiming and Zhang, Xiangyu and Ren, Shaoqing and Sun, Jian},
  booktitle={CVPR},
  pages={770--778},
  year={2016}
}

@article{attention,
  title={Attention is all you need},
  author={Vaswani, Ashish and Shazeer, Noam and Parmar, Niki and Uszkoreit, Jakob and Jones, Llion and Gomez, Aidan N and Kaiser, {\L}ukasz and Polosukhin, Illia},
  journal={NeurIPS},
  volume={30},
  year={2017}
}

@inproceedings{mae,
  title={Masked autoencoders are scalable vision learners},
  author={He, Kaiming and Chen, Xinlei and Xie, Saining and Li, Yanghao and Doll{\'a}r, Piotr and Girshick, Ross},
  booktitle={CVPR},
  pages={16000--16009},
  year={2022}
}

@inproceedings{coco,
  title={Microsoft coco: Common objects in context},
  author={Lin, Tsung-Yi and Maire, Michael and Belongie, Serge and Hays, James and Perona, Pietro and Ramanan, Deva and Doll{\'a}r, Piotr and Zitnick, C Lawrence},
  booktitle={ECCV},
  pages={740--755},
  year={2014},
  organization={Springer}
}

@inproceedings{bib22,
  title={Large-scale datasets for going deeper in image understanding},
  author={Wu, Jiahong and Zheng, He and Zhao, Bo and Li, Yixin and Yan, Baoming and Liang, Rui and Wang, Wenjia and Zhou, Shipei and Lin, Guosen and Fu, Yanwei and others},
  booktitle={ICME},
  pages={1480--1485},
  year={2019},
  organization={IEEE}
}

@inproceedings{yolo,
  title={You only learn once: Universal anatomical landmark detection},
  author={Zhu, Heqin and Yao, Qingsong and Xiao, Li and Zhou, S Kevin},
  booktitle={MICCAI},
  pages={85--95},
  year={2021},
  organization={Springer}
}

@inproceedings{bib41,
  title={Pose recognition with cascade transformers},
  author={Li, Ke and Wang, Shijie and Zhang, Xiang and Xu, Yifan and Xu, Weijian and Tu, Zhuowen},
  booktitle={CVPR},
  pages={1944--1953},
  year={2021}
}

@article{bib42,
  title={A benchmark for comparison of dental radiography analysis algorithms},
  author={Wang, Ching-Wei and Huang, Cheng-Ta and Lee, Jia-Hong and Li, Chung-Hsing and Chang, Sheng-Wei and Siao, Ming-Jhih and Lai, Tat-Ming and Ibragimov, Bulat and Vrtovec, Toma{\v{z}} and Ronneberger, Olaf and others},
  journal={Medical image analysis},
  volume={31},
  pages={63--76},
  year={2016},
  publisher={Elsevier}
}

@article{bib43,
  title={Integrating spatial configuration into heatmap regression based CNNs for landmark localization},
  author={Payer, Christian and {\v{S}}tern, Darko and Bischof, Horst and Urschler, Martin},
  journal={Medical image analysis},
  volume={54},
  pages={207--219},
  year={2019},
  publisher={Elsevier}
}

@inproceedings{bib47,
  title={Batch normalization: Accelerating deep network training by reducing internal covariate shift},
  author={Ioffe, Sergey and Szegedy, Christian},
  booktitle={ICML},
  pages={448--456},
  year={2015},
  organization={pmlr}
}

@article{bib48,
  title={Robust and accurate shape model matching using random forest regression-voting},
  author={Lindner, Claudia and Bromiley, Paul A and Ionita, Mircea C and Cootes, Tim F},
  journal={IEEE transactions on pattern analysis and machine intelligence},
  volume={37},
  number={9},
  pages={1862--1874},
  year={2014},
  publisher={IEEE}
}

@article{bib49,
  title={Integrating geometric configuration and appearance information into a unified framework for anatomical landmark localization},
  author={Urschler, Martin and Ebner, Thomas and {\v{S}}tern, Darko},
  journal={Medical image analysis},
  volume={43},
  pages={23--36},
  year={2018},
  publisher={Elsevier}
}

@article{bib50,
  title={Integrating spatial configuration into heatmap regression based CNNs for landmark localization},
  author={Payer, Christian and {\v{S}}tern, Darko and Bischof, Horst and Urschler, Martin},
  journal={Medical image analysis},
  volume={54},
  pages={207--219},
  year={2019},
  publisher={Elsevier}
}

@inproceedings{unet,
  title={U-net: Convolutional networks for biomedical image segmentation},
  author={Ronneberger, Olaf and Fischer, Philipp and Brox, Thomas},
  booktitle={MICCAI},
  pages={234--241},
  year={2015},
  organization={Springer}
}

@inproceedings{bib52,
  title={From local to global random regression forests: exploring anatomical landmark localization},
  author={{\v{S}}tern, Darko and Ebner, Thomas and Urschler, Martin},
  booktitle={MICCAI},
  pages={221--229},
  year={2016},
  organization={Springer}
}

@article{bib53,
  title={Gaussian error linear units (gelus)},
  author={Hendrycks, Dan and Gimpel, Kevin},
  journal={arXiv},
  year={2016}
}

@article{datr,
  title={DATR: domain-adaptive transformer for multi-domain landmark detection},
  author={Zhu, Heqin and Yao, Qingsong and Zhou, S Kevin},
  journal={arXiv},
  year={2022}
}

@article{jiang2021deep,
  title={Deep learning for COVID-19 chest CT (computed tomography) image analysis: A lesson from lung cancer},
  author={Jiang, Hao and Tang, Shiming and Liu, Weihuang and Zhang, Yang},
  journal={Computational and Structural Biotechnology Journal},
  volume={19},
  pages={1391--1399},
  year={2021},
  publisher={Elsevier}
}

@article{jiang2020geometry,
  title={Geometry-aware cell detection with deep learning},
  author={Jiang, Hao and Li, Sen and Liu, Weihuang and Zheng, Hongjin and Liu, Jinghao and Zhang, Yang},
  journal={Msystems},
  volume={5},
  number={1},
  pages={10--1128},
  year={2020},
  publisher={Am Soc Microbiol}
}

@inproceedings{jia2022visual,
  title={Visual prompt tuning},
  author={Jia, Menglin and Tang, Luming and Chen, Bor-Chun and Cardie, Claire and Belongie, Serge and Hariharan, Bharath and Lim, Ser-Nam},
  booktitle={European Conference on Computer Vision},
  pages={709--727},
  year={2022},
  organization={Springer}
}

@article{zhang2022correction,
  title={Correction of out-of-focus microscopic images by deep learning},
  author={Zhang, Chi and Jiang, Hao and Liu, Weihuang and Li, Junyi and Tang, Shiming and Juhas, Mario and Zhang, Yang},
  journal={Computational and Structural Biotechnology Journal},
  volume={20},
  pages={1957--1966},
  year={2022},
  publisher={Elsevier}
}

@article{liu2020fine,
  title={Fine-grained breast cancer classification with bilinear convolutional neural networks (BCNNs)},
  author={Liu, Weihuang and Juhas, Mario and Zhang, Yang},
  journal={Frontiers in genetics},
  volume={11},
  pages={547327},
  year={2020}
}

@article{liu2024dh,
  title={DH-GAN: Image manipulation localization via a dual homology-aware generative adversarial network},
  author={Liu, Weihuang and Cun, Xiaodong and Pun, Chi-Man},
  journal={Pattern Recognition},
  pages={110658},
  year={2024}
}

@article{liu2023explicit2,
  title={Explicit visual prompting for universal foreground segmentations},
  author={Liu, Weihuang and Shen, Xi and Pun, Chi-Man and Cun, Xiaodong},
  journal={arXiv},
  year={2023}
}

@article{Gong2023GenerativeAF,
  title={Generative AI for brain image computing and brain network computing: a review},
  author={Changwei Gong and Changhong Jing and Xuhang Chen and Chi-Man Pun and Guoli Huang and Ashirbani Saha and Martin Nieuwoudt and Han-Xiong Li and Yong Hu and Shuqiang Wang},
  journal={Frontiers in Neuroscience},
  year={2023},
  volume={17}
}

@inproceedings{liu2024depth,
  title={Depth-aware Test-Time Training for Zero-shot Video Object Segmentation},
  author={Liu, Weihuang and Shen, Xi and Li, Haolun and Bi, Xiuli and Liu, Bo and Pun, Chi-Man and Cun, Xiaodong},
  booktitle={CVPR},
  year={2024}
}

@inproceedings{Huang2023MRIS,
  title={MR Image Super-Resolution Using Wavelet Diffusion for Predicting Alzheimer's Disease},
  author={Guoli Huang and Xuhang Chen and Yanyan Shen and Shuqiang Wang},
  booktitle={BI},
  year={2023}
}

@inproceedings{liu2023coordfill,
  title={Coordfill: Efficient high-resolution image inpainting via parameterized coordinate querying},
  author={Liu, Weihuang and Cun, Xiaodong and Pun, Chi-Man and Xia, Menghan and Zhang, Yong and Wang, Jue},
  booktitle={AAAI},
  year={2023}
}

@inproceedings{Luo2023DevignetHV,
  title={Devignet: High-Resolution Vignetting Removal via a Dual Aggregated Fusion Transformer With Adaptive Channel Expansion},
  author={Shenghong Luo and Xuhang Chen and W. Chen and Zinuo Li and Shuqiang Wang and Chi-Man Pun},
  booktitle={AAAI},
  year={2023}
}

@article{Chen2022ShadocnetLS,
  title={Shadocnet: Learning Spatial-Aware Tokens in Transformer for Document Shadow Removal},
  author={Xuhang Chen and Xiaodong Cun and Chi-Man Pun and Shuqiang Wang},
  journal={ICASSP},
  year={2022},
  pages={1-5}
}

@article{Li2023HighResolutionDS,
  title={High-Resolution Document Shadow Removal via A Large-Scale Real-World Dataset and A Frequency-Aware Shadow Erasing Net},
  author={Zinuo Li and Xuhang Chen and Chi-Man Pun and Xiaodong Cun},
  journal={ICCV},
  year={2023},
  pages={12415-12424},
}

@inproceedings{Chen2023ALF,
  author = {Li, Zinuo and Chen, Xuhang and Wang, Shuqiang and Pun, Chi-Man},
  booktitle = {IJCAI},
  pages = {1160-1168},
  title = {A Large-Scale Film Style Dataset for Learning Multi-frequency Driven Film Enhancement},
  year = {2023},
}

@inproceedings{Chen2024-dg,
  author = {Chen, Xuhang and Lei, Baiying and Pun, Chi-Man and Wang, Shuqiang},
  booktitle = {PRCV},
  pages = {16-26},
  title = {Brain Diffuser: An End-to-End Brain Image to Brain Network Pipeline},
  year = {2023},
}

@article{Chen2023MedPromptCP,
  title={MedPrompt: Cross-Modal Prompting for Multi-Task Medical Image Translation},
  author={Xuhang Chen and Chi-Man Pun and Shuqiang Wang},
  journal={ArXiv},
  year={2023}
}

@article{li2022wavenhancer,
    title = {WavEnhancer: Unifying Wavelet and Transformer for Image Enhancement},
    journal = {Journal of Computer Science and Technology},
    volume = {39},
    number = {2},
    pages = {336-345},
    year = {2024},
    author = {Zinuo Li and Xuhang Chen and Shuna Guo and Shuqiang Wang and Chi-Man Pun}
}

@inproceedings{chen7,
    title={Dual-Hybrid Attention Network for Specular Highlight Removal},
    author={Xiaojiao Guo and Xuhang Chen and Shenghong Luo and Shuqiang Wang and Chi-Man Pun},
    booktitle={ACM MM},
    year={2024}
}

@article{chen10,
    title={MFDNet: Multi-Frequency Deflare Network for efficient nighttime flare removal},
    author={Jiang, Yiguo and Chen, Xuhang and Pun, Chi-Man and Wang, Shuqiang and Feng, Wei},
    journal={The Visual Computer},
    pages={1--14},
    year={2024}
}

@inproceedings{chen11,
    title={Generative ai enables eeg data augmentation for alzheimer’s disease detection via diffusion model},
    author={Zhou, Tong and Chen, Xuhang and Shen, Yanyan and Nieuwoudt, Martin and Pun, Chi-Man and Wang, Shuqiang},
    booktitle={ISPCE-ASIA},
    pages={1--6},
    year={2023}
}

@article{chen12,
    title={Weakly supervised semantic segmentation via saliency perception with uncertainty-guided noise suppression},
    author={Liu, Xinyi and Huang, Guoheng and Yuan, Xiaochen and Zheng, Zewen and Zhong, Guo and Chen, Xuhang and Pun, Chi-Man},
    journal={The Visual Computer},
    pages={1--16},
    year={2024}
}

@article{chen13,
    title={Psanet: prototype-guided salient attention for few-shot segmentation},
    author={Li, Hao and Huang, Guoheng and Yuan, Xiaochen and Zheng, Zewen and Chen, Xuhang and Zhong, Guo and Pun, Chi-Man},
    journal={The Visual Computer},
    pages={1--15},
    year={2024}
}

@article{chen14,
  title={RM-UNet: UNet-like Mamba with rotational SSM module for medical image segmentation},
  author={Hao Tang and Guoheng Huang and Lianglun Cheng and Xiaochen Yuan and Qi Tao and Xuhang Chen and Guo Zhong and Xiaohui Yang},
  journal={Signal, Image and Video Processing},
  year={2024}
}

@article{chen15,
	title        = {DocDeshadower: Frequency-aware Transformer for Document Shadow Removal},
	author       = {Ziyang Zhou and Yingtie Lei and Xuhang Chen and Shenghong Luo and Wenjun Zhang and Chi-Man Pun and Zhen Wang},
	year         = 2023,
	journal      = {arXiv}
}

@article{zhang1,
  title={Sienet: Siamese expansion network for image extrapolation},
  author={Zhang, Xiaofeng and Chen, Feng and Wang, Cailing and Tao, Ming and Jiang, Guo-Ping},
  journal={IEEE Signal Processing Letters},
  volume={27},
  pages={1590--1594},
  year={2020},
  publisher={IEEE}
}

@inproceedings{zhang2,
  title={SpA-Former: An Effective and lightweight Transformer for image shadow removal},
  author={Zhang, Xiaofeng and Zhao, Yudi and Gu, Chaochen and Lu, Changsheng and Zhu, Shanying},
  booktitle={2023 International Joint Conference on Neural Networks (IJCNN)},
  pages={1--8},
  year={2023},
  organization={IEEE}
}

@article{zhang3,
  title={Memory augment is All You Need for image restoration},
  author={Zhang, Xiao Feng and Gu, Chao Chen and Zhu, Shan Ying},
  journal={arXiv preprint arXiv:2309.01377},
  year={2023}
}

@article{zhang4,
  title={Enlighten-anything: When segment anything model meets low-light image enhancement},
  author={Zhao, Qihan and Zhang, Xiaofeng and Tang, Hao and Gu, Chaochen and Zhu, Shanying},
  journal={arXiv preprint arXiv:2306.10286},
  year={2023}
}

@article{zhang5,
  title={MuralDiff: Diffusion for Ancient Murals Restoration on Large-Scale Pre-Training},
  author={Xu, Zishan and Zhang, Xiaofeng and Chen, Wei and Liu, Jueting and Xu, Tingting and Wang, Zehua},
  journal={IEEE Transactions on Emerging Topics in Computational Intelligence},
  year={2024},
  publisher={IEEE}
}

@article{zhang6,
  title={Shadclips: When Parameter-Efficient Fine-Tuning with Multimodal Meets Shadow Removal},
  author={Zhang, Xiaofeng and Xu, Zishan and Tang, Hao and Gu, Chaochen and Zhu, Shanying and Guan, Xinping},
  year={2024}
}

@article{zhang7,
  title={From Redundancy to Relevance: Enhancing Explainability in Multimodal Large Language Models},
  author={Zhang, Xiaofeng and Shen, Chen and Yuan, Xiaosong and Yan, Shaotian and Xie, Liang and Wang, Wenxiao and Gu, Chaochen and Tang, Hao and Ye, Jieping},
  journal={arXiv preprint arXiv:2406.06579},
  year={2024}
}

@article{zhang8,
  title={DOPRA: Decoding Over-accumulation Penalization and Re-allocation in Specific Weighting Layer},
  author={Wei, Jinfeng and Zhang, Xiaofeng},
  journal={arXiv preprint arXiv:2407.15130},
  year={2024}
}

@article{zhang9,
  title={AFFNet: attention mechanism network based on fusion feature for image cloud removal},
  author={Shen, Runhan and Zhang, Xiaofeng and Xiang, Yonggang},
  journal={International Journal of Pattern Recognition and Artificial Intelligence},
  volume={36},
  number={08},
  pages={2254014},
  year={2022},
  publisher={World Scientific}
}

@article{zhang10,
  title={LL-Diff: Low-light image enhancement utilizing Langevin sampling diffusion},
  author={Ding, Boren and Zhang, Xiaofeng and Yu, Zekun and Zhao, Chuangxin and Yao, JiaWei and Hui, Zheng},
  journal={International Journal of Pattern Recognition and Artificial Intelligence},
  year={2024},
  publisher={World Scientific}
}

@article{li1,
  title={Complementation-reinforced network for integrated reconstruction and segmentation of pulmonary gas MRI with high acceleration},
  author={Li, Zimeng and Xiao, Sa and Wang, Cheng and Li, Haidong and Zhao, Xiuchao and Zhou, Qian and Rao, Qiuchen and Fang, Yuan and Xie, Junshuai and Shi, Lei and others},
  journal={Medical Physics},
  volume={51},
  number={1},
  pages={378--393},
  year={2024}
}

@article{li2,
  title={Encoding enhanced complex CNN for accurate and highly accelerated MRI},
  author={Li, Zimeng and Xiao, Sa and Wang, Cheng and Li, Haidong and Zhao, Xiuchao and Duan, Caohui and Zhou, Qian and Rao, Qiuchen and Fang, Yuan and Xie, Junshuai and others},
  journal={IEEE Transactions on Medical Imaging},
  year={2024}
}

@inproceedings{li2020diagnosis,
  title={Diagnosis of Parkinson's Disease with a hybrid feature selection algorithm based on a discrete artificial bee colony},
  author={Li, Haolun and Zong, Rui and Xu, Xin and Pan, Long Sheng and Dai, Qionghai and Xu, Feng and Gao, Hao and Wang, Wensheng},
  booktitle={Medical Imaging 2020: Computer-Aided Diagnosis},
  volume={11314},
  pages={640--647},
  year={2020},
  organization={SPIE}
}

@article{li2021hybrid,
  title={A hybrid feature selection algorithm based on a discrete artificial bee colony for Parkinson's diagnosis},
  author={Li, Haolun and Pun, Chi-Man and Xu, Feng and Pan, Longsheng and Zong, Rui and Gao, Hao and Lu, Huimin},
  journal={ACM Transactions on Internet Technology},
  volume={21},
  number={3},
  pages={1--22},
  year={2021},
  publisher={ACM New York, NY}
}

@inproceedings{li2023optimized,
  title={An optimized-skeleton-based Parkinsonian gait auxiliary diagnosis method with both monitoring indicators and assisted ratings},
  author={Li, Gaoqi and Pun, Chi-Man and Li, Haolun and Xiong, Jian and Xu, Feng and Gao, Hao},
  booktitle={2023 IEEE International Conference on Bioinformatics and Biomedicine (BIBM)},
  pages={2011--2016},
  year={2023},
  organization={IEEE}
}

@inproceedings{li2023cee,
  title={Cee-net: complementary end-to-end network for 3d human pose generation and estimation},
  author={Li, Haolun and Pun, Chi-Man},
  booktitle={Proceedings of the AAAI Conference on Artificial Intelligence},
  volume={37},
  number={1},
  pages={1305--1313},
  year={2023}
}

@article{li2022monocular,
  title={Monocular robust 3d human localization by global and body-parts depth awareness},
  author={Li, Haolun and Pun, Chi-Man},
  journal={IEEE Transactions on Circuits and Systems for Video Technology},
  volume={32},
  number={11},
  pages={7692--7705},
  year={2022},
  publisher={IEEE}
}

@inproceedings{song2024local,
  title={Local Optimization Networks for Multi-View Multi-Person Human Posture Estimation},
  author={Song, Jucheng and Pun, Chi-Man and Li, Haolun and Lan, Rushi and Xie, Jiu-Cheng and Gao, Hao},
  booktitle={ICASSP 2024-2024 IEEE International Conference on Acoustics, Speech and Signal Processing (ICASSP)},
  pages={3995--3999},
  year={2024},
  organization={IEEE}
}

@article{li2022few,
  title={Few-shot object detection via high-and-low resolution representation},
  author={Li, Haolun and Ge, Senlin and Gao, Chuyi and Gao, Hao},
  journal={Computers and Electrical Engineering},
  volume={104},
  pages={108438},
  year={2022},
  publisher={Elsevier}
}

@inproceedings{huang2024deformmlp,
  title={DeformMLP: Dynamic Large-Scale Receptive Field MLP Networks for Human Motion Prediction},
  author={Huang, Haitao and Pun, Chi-Man and Li, Haolun and Liu, Mengqi and Xiong, Jian and Gao, Hao},
  booktitle={ICASSP 2024-2024 IEEE International Conference on Acoustics, Speech and Signal Processing (ICASSP)},
  pages={5200--5204},
  year={2024},
  organization={IEEE}
}

@article{yang2024adaptive,
  title={Adaptive Spatial-Temporal Graph-Mixer for Human Motion Prediction},
  author={Yang, Shubo and Li, Haolun and Pun, Chi-Man and Du, Chun and Gao, Hao},
  journal={IEEE Signal Processing Letters},
  year={2024},
  publisher={IEEE}
}

\end{document}